\documentclass[letterpaper]{article} 
\usepackage[preprint]{aaai2027}  
\usepackage[hyphens]{url}  
\usepackage{graphicx} 
\usepackage{natbib}  
\usepackage{caption} 
\usepackage{newfloat}
\usepackage{listings}

\usepackage{booktabs}
\usepackage{makecell}
\usepackage{enumitem}
\usepackage{multirow}
\usepackage{subcaption}
\usepackage{amssymb}
\usepackage{makecell}
\usepackage{bm}

\newcommand{\q}[1]{\bf{Q#1}}

\title{Cross-Regional Grapevine Cold Hardiness Prediction via Learned Multimodal Latent Representations}

\author {
     William Solow\textsuperscript{\rm 1},
     Paola Pesantez-Cabrera\textsuperscript{\rm 2},
     Markus Keller\textsuperscript{\rm 2},
     Lav Khot\textsuperscript{\rm 2},
     Sandhya Saisubramanian\textsuperscript{\rm 1},
     Alan Fern\textsuperscript{\rm 1}
 }
 \affiliations {
     \textsuperscript{\rm 1}Oregon State University, 
     \textsuperscript{\rm 2}Washington State University\\
     \{soloww,sandhya.sai,afern\}@oregonstate.edu,
     \{p.pesantezcabrera,mkeller,lav.khot\}@wsu.edu
 }

\begin{document}

\maketitle
  
\begin{abstract}
Accurate daily predictions of cold hardiness in woody plants are critical in regions where freezing temperatures can damage dormant buds and reduce seasonal yield. Existing biophysical, hybrid, and deep learning models have shown high predictive accuracy when trained on local data but remain largely \emph{site-specific}. The limited availability of cold hardiness data, coupled with the lack of principled methods for transferring cold hardiness predictions to new regions and cultivars, has limited the broader adoption and practical utility of these approaches, particularly in data-scarce regions. To address these limitations, we propose a cold hardiness prediction framework that learns a transferable latent representation by capturing region-specific variation through learned embeddings. To enable prediction in previously unseen regions, we infer embeddings from (1) text descriptions of the cultivar and growing region, and (2) limited historical observations, supporting both zero-shot and few-shot transfer. Experiments on datasets from six regions across North America demonstrate that our approach consistently outperforms state-of-the-art cold hardiness prediction methods, yielding more accurate predictions and substantially improving transfer to data-scarce regions.

\end{abstract}

\section{Introduction}
\begin{figure}[t]
    \centering
    \begin{subfigure}{0.495\linewidth}
        \centering
        \includegraphics[width=\textwidth]{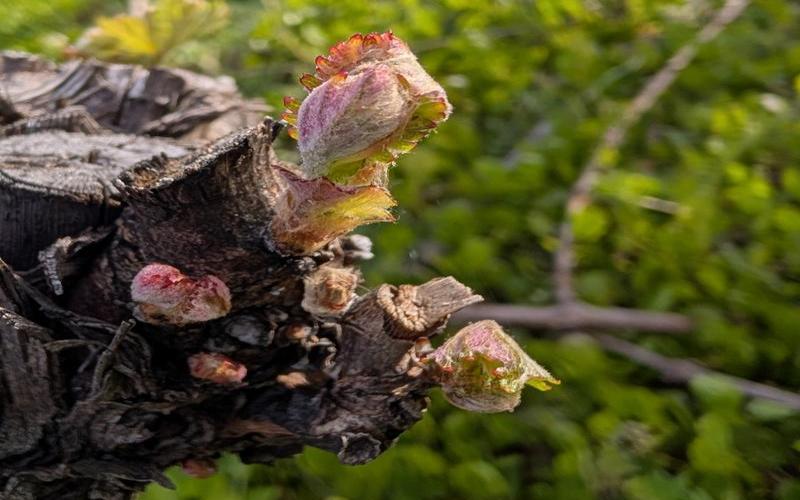}
    \end{subfigure}
    \hfill
    \begin{subfigure}{0.495\linewidth}
        \centering
        \includegraphics[width=\textwidth]{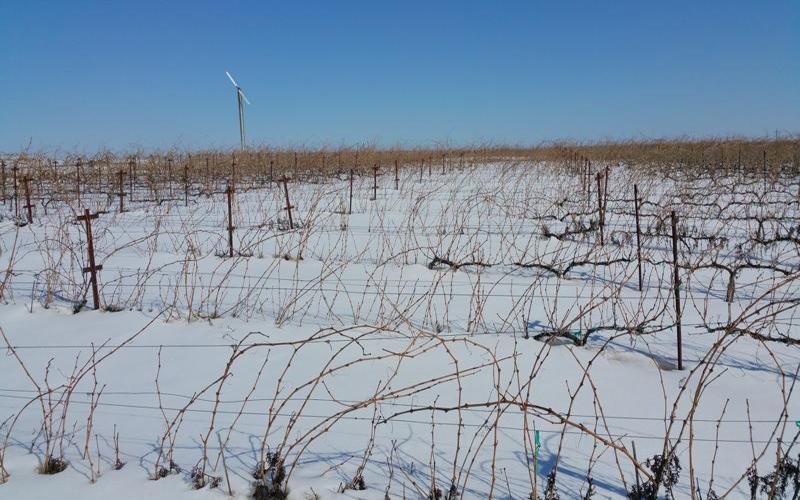}
    \end{subfigure}

    \begin{subfigure}{0.495\linewidth}
        \centering
        \includegraphics[width=\textwidth]{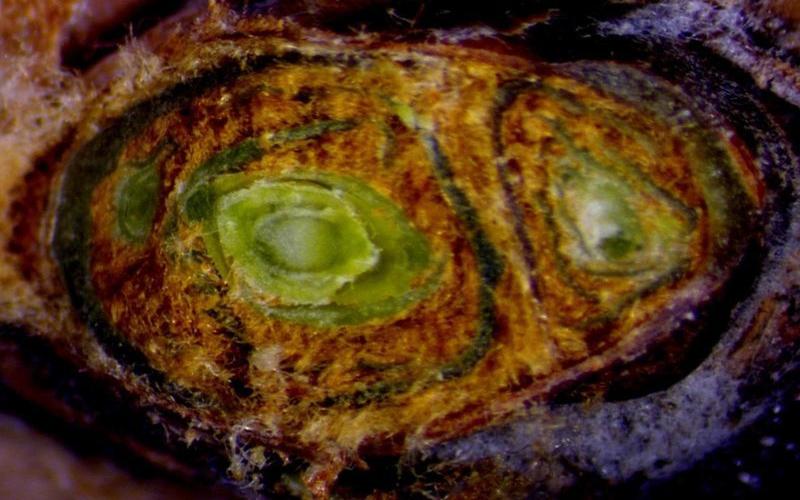}
    \end{subfigure}
    \hfill
    \begin{subfigure}{0.495\linewidth}
        \centering
        \includegraphics[width=\textwidth]{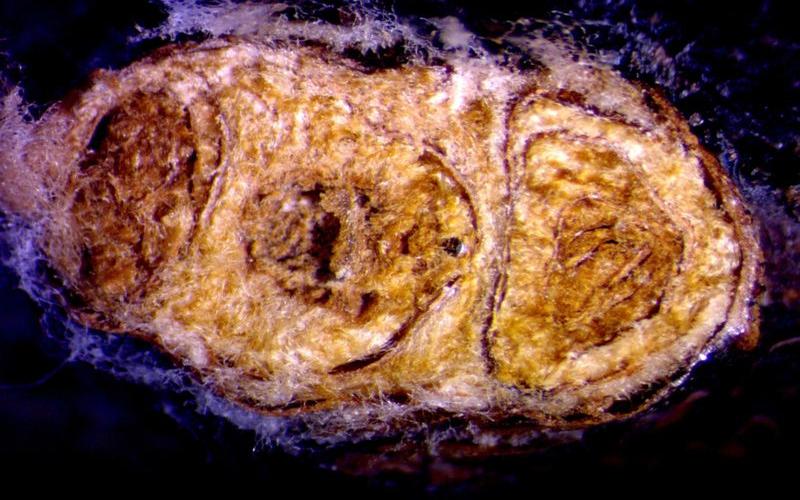}
    \end{subfigure}

    \caption{Top: (Left) A grapevine bud emerging from dormancy when it is most susceptible to damaging frost events. (Right) The vineyard in Prosser, WA where our collaborators collect bud samples; harsh winter environments necessitate careful management decisions to ensure dormant buds survive. Bottom: The cross sections of a healthy (left) and damaged (right) grapevine bud during the dormancy season.} 
    \label{fig:collage}
\end{figure}

Viticulture in North American regions with cold winters (Washington State, British Columbia's Okanagan Valley, Ontario's Niagara Peninsula, New York State) accounts for over \$30 billion dollars in yearly economic impact~\citep{mari2025}. A major challenge in these cool-climate regions is the accurate prediction of grapevine cold hardiness, or the lethal temperature at which 50\% of dormant buds freeze (LT50). In the advent of an unusually cold event (Figure~\ref{fig:collage}, top right) and insufficient vine cold hardiness, growers may deploy preventive measures such as wind fans or kerosene heaters to warm the air in their vineyards to help mitigate substantial economic loss~\citep{jones2000}. Given the significant cost of deploying these preventive measures, they should only be used when absolutely necessary, underscoring the need for accurate LT50 predictions for precision management.

Throughout the winter, grapevine cold hardiness varies with the weather, meaning that it must be measured frequently for informed vineyard decision making~\citep{pagter2013}. However, LT50 measurements require specialized lab equipment and trained technicians~\citep{mills2006}, and cold damage can only be assessed after removal of buds from the vine (Figure~\ref{fig:collage}, bottom). As a result, it is infeasible for growers to monitor the cold hardiness of their grapevines in real time. This motivated the development of predictive models to serve as decision support tools for growers~\citep{ferguson2014,saxena2023a,wang2024a}. Given the extensive effort required to collect sufficient per-cultivar data, curating diverse multi-region datasets is challenging. Consequently, existing modeling approaches rely on site- or cultivar-specific data, leading to poor out-of-region generalization~\citep{rubio2020}. As some cool-climate regions (e.g. Michigan, Nova Scotia) contain insufficient data for model calibration, under-informed vineyard management decisions are common and affect seasonal yield and quality. 

To make actionable predictions in data-scarce regions, a naive approach is to directly transfer predictions from regions with more data. However, inter-region prediction transfer is challenging because the cold hardiness response of the same cultivar varies substantially across regions due to unobserved latent factors such as soil properties, local climate, and management practices~\citep{khanizadeh2005}. We address this challenge by training a multi-task model across the cross product of regions, weather stations, vineyards, and cultivars, to learn latent representations that encode information about region-level cold hardiness trends to enable knowledge transfer across geographically distinct regions. To enable transfer of predictions to previously unseen regions and cultivars, we introduce two complementary mechanisms that infer latent representations from (i) textual descriptions of the cultivar and growing region and (ii) limited historical observations, thus supporting both zero-shot and few-shot transfer. Our framework is applicable to biophysical, deep learning, and hybrid cold hardiness models, allowing a unified comparison of transfer performance across these modeling paradigms.

\paragraph{Scope, Contributions, and Track Relevance} This work is motivated by the operational needs of our primary stakeholders: grapevine growers and regional agricultural organizations in the Pacific Northwest (PNW), USA, who face increasing risks from weather variability and climate stress~\citep{reynolds2022, keller2010a}. Our contributions are three-fold: we (1) present a multi-task cold hardiness modeling framework that unifies biophysical, deep learning, and hybrid approaches, enabling systematic comparison of predictive accuracy, biological realism, and transferability; (2) investigate two methods to enable cold hardiness prediction transfer to previously unseen regions and cultivars by inferring latent representations from (a) text descriptions and (b) limited historical observations; and (3) evaluate the proposed approach on six real world datasets spanning cool-climate regions in North America. Our cold hardiness model has been \textit{deployed} on AgWeatherNet for the 2026-2027 dormancy season and we are actively exploring potential deployment of our transfer learning approach with our partners in the PNW region. 

\section{Related Work}

\subsubsection{Cold Hardiness Modeling}
Grapevine cold hardiness is the property of dormant grapevine buds to tolerate freezing temperatures without lethal tissue damage~\citep{browse2001}. Cold hardiness changes throughout the dormancy season in response to ambient temperatures and other latent variables~\citep{gusta2013}. Given the importance of grapevine cold hardiness for vineyard site selection, cultivar choice, and preventative frost management, extensive research has gone into the collection of grapevine cold hardiness data using differential thermal analysis~\citep{mills2006}. The Ferguson model~\citep{ferguson2014} is the premier biophysical model for predicting grapevine cold hardiness and is currently hosted on AgWeatherNet~\citep{c:agnet} where it is available as a decision support tool for grapevine growers. 

To address the poor performance of the Ferguson model on cultivars with few seasons of historical data,~\citet{saxena2023a} proposed GrapeHardiNet, a deep learning model that leverages multi-task learning for efficient data sharing across cultivars. Both the Ferguson and GrapeHardiNet models perform site- and cultivar- specific predictions; in contrast, NYUS2~\citep{wang2024a} aggregates data across diverse viticultural regions with per-cultivar predictions, but does not account for regional variance. Noting the prevalence of biologically inconsistent predictions leading up to bud break,~\citet{solow2026} proposed a hybrid modeling approach to modulate the parameters of biophysical models in response to weather. 
However, these approaches do not support transfer as they rely on the availability of local training data to produce LT50 predictions.

\vspace{3pt}
\noindent \textbf{Generalizing LT50 Predictions Across Regions}
GrapeHardiNet~\citep{saxena2023a} proposed a multi-tasking approach for LT50 predictions that aggregates data across cultivars at a \emph{specific site}, assuming they experience identical weather and soil properties, and have their LT50 measured from the same lab. While this assumption is reasonable for site-specific aggregation, it is violated when aggregating \emph{across regions}, necessitating an approach that considers regional variance~\citep{rubio2020}. We address this limitation through multi-task and transfer learning~\citep{zhuang2021, caruana1997}. 
To the best of our knowledge, this is the first study of transfer learning for LT50 prediction.

\begin{figure*}[t]
    \centering
    \includegraphics[width=0.89\linewidth]{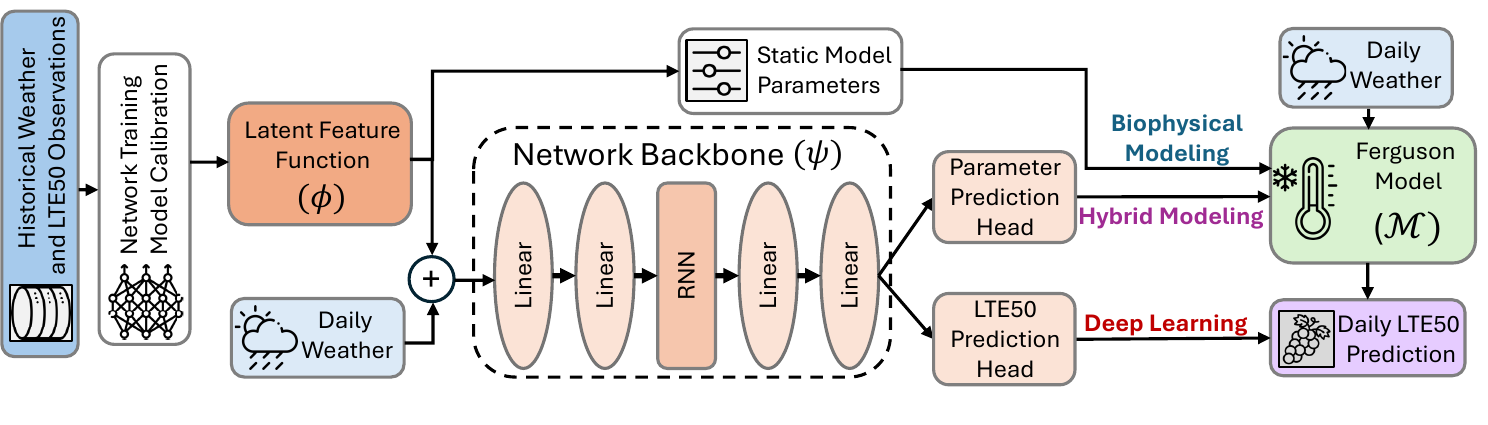}
    \caption{A unifying overview of the three primary modeling paradigms for cold hardiness prediction. The top, middle, and bottom represent the pipelines for biophysical model calibration, hybrid modeling, and deep learning, respectively. All result in daily LT50 predictions across cultivars. Brown and blue colors represent neural components and data within our implementation.  }
    \label{fig:unifying_framework}
\end{figure*}

\section{Problem Formulation}
The cold hardiness prediction problem can be viewed as a multi-task sequence prediction problem, where cold hardiness is an unobserved latent function $\mathcal{F}$ mapping daily weather, regions, weather stations, vineyards, and cultivars to LT50 values. Each prediction task $\tau\in\mathcal{T}$ corresponds to a unique combination of region, weather station, vineyard, and cultivar. Although these tasks share the same underlying biological processes governing cold hardiness, they differ in their environmental conditions and cultivar-specific characteristics. Learning across related tasks therefore provides an opportunity to improve prediction accuracy and facilitate transfer to new regions and cultivars. 

Let $R$, $S$, $V$, and $C$ denote the sets of regions, weather stations, vineyards, and cultivars. Each task is defined as $\tau = (r,s,v,c) \in R \times S \times V \times C$. Let $W\subseteq\mathbb{R}^m$ be the set of daily weather feature vectors. LT50 datasets are collected seasonally from September 7th to May 15th each year. A sequence of data for season $k$ is denoted by $X^k_{r,s,v,c}=(w_1,y_1,\ldots,w_H,y_H)$ where $w_i$ is the daily weather and $y_i$ is the LT50 observation, if one exists. Given a dataset $D$ of LT50 and weather observations $w_{1:H}$, our objective is to produce a model $\widehat{\mathcal{F}}(w_{1:H},\tau)$ that maps a sequence of weather observations, together with the task identity, to a sequence of LT50 predictions $(\hat{y}_1, \hat{y}_2,\ldots,\hat{y}_H)$, while sharing information across tasks to improve prediction accuracy and enable transfer to previously unseen regions or cultivars.



Our multi-task framework for cold hardiness prediction transfer is composed of three components: (1) a task encoder function $\phi$ that captures latent information about the prediction task; (2) a function $\psi$ that encodes temporal relationships between weather and crop state, conditioned on the latent state; and (3) a biophysical model $\mathcal{M}$ that maps weather observations to LT50 values. Formally, 

\begin{itemize}
    \item $\phi:R\times S\times V\times C\to\mathbb{R}^n$ is a mapping from the regions, weather stations, vineyards and cultivars to a latent feature space which encodes information about the specific prediction task;
    \item  $\psi:\mathbb{R}^{n+m}\to\mathbb{R}^p$ is a mapping from the cross product of the latent feature space and weather space to a $p$-dimensional vector in $\mathbb{R}^p$; and
    \item  $\mathcal{M}:\mathbb{R}^{p+m}\to\mathbb{R}$ is a biophysical model, mapping the cross product of the parameter space and weather space to the output space, denoting the LT50 prediction. 
\end{itemize} 

Composing the three functions together, the multi-task prediction model is $\hat{y_i} = \mathcal{M}[\psi[\phi(r,s,v,c),w_i],w_i]$ (Figure~\ref{fig:unifying_framework}). This model can be trained end-to-end, resulting in learned latent representations denoted by $\phi$, for each task $\tau$, which we leverage for our transfer approach.

\paragraph{Unifying Modeling Lens}
Existing approaches to LT50 prediction approximate $\mathcal{F}$ through three broad modeling paradigms: biophysical, deep learning, and hybrid models. These paradigms have largely emerged from separate research communities, with biophysical models originating in agricultural science and deep learning and hybrid models in artificial intelligence (AI), resulting in complementary strengths but few common abstractions for systematic comparison. While they all approximate $\mathcal{F}$, their primary differences lie in how they represent task-specific information and model the relationship between weather observations and LT50. We show that these seemingly disparate approaches can be expressed as instantiations of our multi-task formalism, enabling systematic comparison between them and incorporating transferable latent representations (Fig.~\ref{fig:unifying_framework}). 
 
\emph{Biophysical models} parameterize a mechanistic model using the latent representation by setting $n=p$, using the parameters as the latent representation function $\phi$, and letting $\psi$ be the identity function: $\hat{y_i} = \mathcal{M}[\phi(r,s,v,c),w_i]$. \emph{Deep learning models} map the latent representation and weather observations to LT50 by letting $\mathcal{M}$ be the identity function and $p=1$, bypassing the structure provided by the biophysical model: $\hat{y_i}=\psi[\phi(r,s,v,c),w_i]$. \emph{Hybrid models} retain all components, with $\psi$ predicting the parameters of the biophysical model $\mathcal{M}$: $\hat{y_i} = \mathcal{M}[\psi[\phi(r,s,v,c),w_i],w_i]$. 

While straightforward, this unified view enables, for the first time, systematic comparison between these modeling paradigms and understanding their trade-offs. Biophysical models produce biologically realistic output, but struggle to capture non-linear temporal dependencies conditioned on the weather~\citep{badeck2004} without an expressive function approximator $\psi$. Deep learning approaches provide greater representational flexibility but lack the structural constraints, imposed by a biophysical model $\mathcal{M}$, needed to ensure biologically consistent predictions. Hybrid models combine the strengths of both approaches by using a neural network as $\psi$ to calibrate the parameters of the biophysical model $\mathcal{M}$, based on the learned latent representation $\phi$ and weather history. This enables data-driven adaptation to produce accurate and biologically consistent predictions.

\section{Transfer Learning of LT50 Predictions}
Given a cold hardiness model $\widehat{\mathcal{F}}$, a simple transfer strategy to predict the LT50 of cultivar $c$ in a new region is to average the LT50 predictions for $c$ across all regions in the training set. However, this approach discounts regional variation and does not leverage available observations or priors of the target $c$. Within our framework, predictions are conditioned on the learned latent vector $\phi(r,s,v,c)$ for each region, weather station, vineyard and cultivar. While these base embeddings can be learned for entities observed during training, they are unavailable for unseen regions or cultivars encountered during deployment, resulting in poor predictions when used. To address this challenge, we learn approximations of the task, $\phi$, using textual descriptions or limited historical data.

\subsubsection{How Learned Latent Features Inform Transfer} 
\begin{figure}[t]
    \centering
    \includegraphics[width=0.85\linewidth]{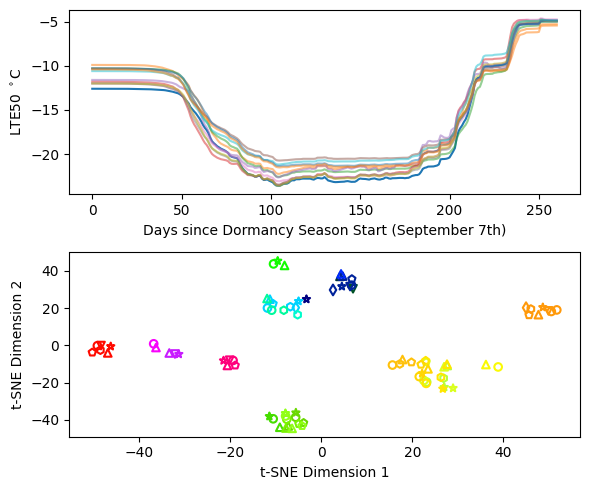}
    \caption{\emph{Top}: LT50 predictions for 12 cultivars at a vineyard in the BCOV region. By varying $\phi$, the predictions from $\psi$ vary substantially. \textit{Bottom}: Dimensionality reduction on the output of $\phi$ using t-SNE. Colors represent different vineyards across six regions and shapes represent the 12 cultivars.}
    \label{fig:embedding_effect}
\end{figure}

To motivate our transfer learning approach, we first examine the latent task representations learned by $\phi$ in our multi-task model $\widehat{\mathcal{F}}$. Figure~\ref{fig:embedding_effect} (top) illustrates the effect of varying the learned representation while holding the weather observations fixed. For 12 cultivars, each associated with a different latent representation, the same model $\widehat{\mathcal{F}}$ trained on two regions, produces substantially different predictions for a single region. This indicates that the latent representation captures the factors governing cold hardiness response beyond the observed weather inputs and is therefore the critical quantity for transferring predictions to unseen regions and cultivars.

We further examine the structure of the learned latent space using t-SNE (Figure~\ref{fig:embedding_effect}, bottom). The embeddings cluster primarily by region rather than cultivar, suggesting that regional characteristics have a stronger influence on cold hardiness than cultivar identity alone (vineyards in the same region are a similar color). This follows the empirical observation that LT50 varies by region~\citep{rubio2020} but challenges the cultivar-specific modeling paradigm~\citep{ferguson2014}, making it clear that growers need specific models calibrated to their vineyard or region. We expand on this discussion in \textit{Appendix E}. 

\begin{figure}
    \centering
    \includegraphics[width=0.95\linewidth]{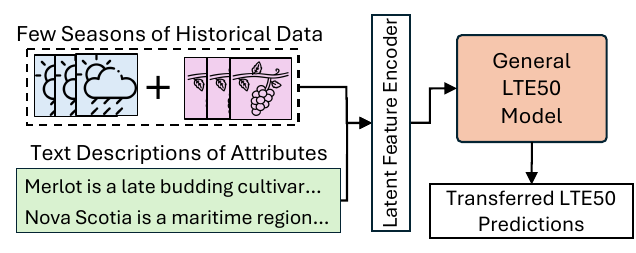}
    \caption{Our proposed approach for transferring LT50 predictions. Past observations and text descriptions are encoded to approximate latent representation $\phi$ that is used by the LT50 model to make predictions in unseen regions.}
    \label{fig:transfer_methods}
\end{figure}

Motivated by these observations, we develop two transfer learning approaches that approximate $\phi$ for unseen regions and cultivars in both zero-shot and few-shot settings. 
Our proposed methods approximate $\phi$ to elicit a wide variety of cold hardiness responses in $\widehat{\mathcal{F}}$ \emph{without} model retraining for prediction of new tasks, e.g. at new vineyards. We detail both approaches below and provide an overview in Figure~\ref{fig:transfer_methods}.
 
\subsubsection{Transfer via Text Embeddings} Inspired by prior works that use text embedding to augment latent spaces for information retrieval~\cite{gao2021}, we hypothesize that text descriptions can be used to encode useful information about the cold hardiness properties of cultivars across regions. We use a text description for each cultivar, region etc. Below are two examples of the text descriptions used in our experiments for the region and cultivar, created in consultation with domain experts: 

\noindent \textbf{Region example:} \emph{Washington State is a semi-arid, continental-influenced wine region with hot, sunny days, cool nights, and low annual precipitation in the rain-shadow of the Cascade Mountains, where well-drained alluvial and volcanic soils combined with irrigation allow precise control of vine vigor and high-quality vinifera grape production.}

\noindent \textbf{Cultivar example:} \emph{Cabernet Franc is an early- to mid-budding, mid-ripening red grape cultivar with medium vigor and loose to moderately compact clusters, producing wines with moderate acidity and a lighter, translucent red color. It has good cold-hardiness (tolerating around -15C/5F) and prefers well-drained soils such as gravelly loam or limestone.}

Such text descriptions can be drafted in consultation with domain experts or from existing scientific literature that describes these cultivars and regions (additional examples are in \textit{Appendix D}). We use the E5 sentence embedding model~\cite{wang2024multilingual} to encode relevant features from sentence descriptions of cultivars, vineyards, weather stations, and regions to produce 4096 dimensional embeddings. We down-project each of the embeddings from the E5 model for the region, weather station, vineyard, and cultivar to size $n/4$ using a multi-layer perception and concatenate them to obtain a vector of size $n$ that approximates the latent representation of the task, $\phi$. As text descriptions can be created for any region, weather station, vineyard, or cultivar, this approximation of $\phi$ can be used in our framework to enable zero-shot transfer to previously unseen locations or cultivars.

\paragraph{Transfer Using Limited Data}
Text embeddings are appropriate when no prior data is available, but in some cases growers may have a few seasons of LT50 observations with accompanying weather observations. While this data is insufficient to train an accurate predictive model, it can be leveraged to learn useful latent features for the cultivar and region where the data were collected. This setting is inherently challenging given that LT50 observations are irregular within a daily weather sequence, with measurements often made weekly, biweekly or monthly~\cite{mills2006}, inconsistently between years. As the LT50 observations are conditioned on the daily weather, cultivar, and region, the latent representation must capture both the daily weather and observed LT50 (more details in \textit{Appendix D)}. 

Any approach that maps two sequences to a latent space can be used to approximate $\phi$ from sparse and irregular observations. We use a cross attention mechanism~\citep{vaswani2017} to perform sequence-to-sequence alignment between the LT50 ($5\leq L\leq20$) and weather observations ($T=261$) where $L$ and $T$ are the length of each sequence respectively, given the ability of transformer architecture to learn distant temporal relations which is a defining feature of LT50 datasets. Our method is as follows: we append a day of season feature to each LT50 observation, as they are not uniformly distributed throughout the sequence. We then pass both sequences through GRUs~\cite{chung2014} with a hidden dimension of $K=128$ to obtain a latent representation of each sequence. After encoding, the weather data is a $T\times K$ matrix and the LT50 data is a $L\times K$ matrix. We then use a cross attention head such that the $T$ length weather sequence attends over the $L$ length LT50 sequence to obtain a latent $T\times K$ dimensional matrix that represents both the weather and cold hardiness for the given season. Using softmax pooling over the $T$ length sequence, we obtain a $K$ dimensional feature vector. When there is more than one season of data present, we repeat the process for each season and average the result. The resulting feature vector is down-projected to a latent vector of size $n$ for use in model $\widehat{\mathcal{F}}$ resulting in a method for few-shot transfer.

\begin{table*}[t]
\centering
\setlength{\tabcolsep}{3mm}
{\fontsize{9}{09}\selectfont
\begin{tabular}{l|rrr|rr}
\toprule
Regions     & Ferguson      & NYUS2.2 & GrapeHardiNet & DL Model & Hybrid Model          \\
     \midrule
BCOV & 1.53 $\pm$ 0.15 & 1.52 $\pm$ 0.42 & 1.31 $\pm$ 0.48 & \bf{1.11 $\pm$ 0.38}$^*$ & 1.19 $\pm$ 0.37 \\ 
ONNP & 1.39 $\pm$ 0.11 & 2.13 $\pm$ 0.60 & 1.19 $\pm$ 0.34 & \bf{0.99 $\pm$ 0.29}$^*$ & 1.03 $\pm$ 0.31 \\
WA   & 1.78 $\pm$ 0.12 & 2.06 $\pm$ 0.47 & 1.30 $\pm$ 0.34 & \bf{1.22 $\pm$ 0.25}$^*$ & 1.23 $\pm$ 0.21 \\
NY   & 2.78 $\pm$ 0.56 & 1.86 $\pm$ 0.54 & 1.68 $\pm$ 0.37 & \bf{1.52 $\pm$ 0.39}$^*$ & 1.63 $\pm$ 0.50 \\
NS   & 2.59 $\pm$ 0.55 & 1.87 $\pm$ 0.58 & 1.55 $\pm$ 0.26 & \bf{1.47 $\pm$ 0.26}\phantom{$^*$} & 1.53 $\pm$ 0.27 \\
MI   & 3.67 $\pm$ 1.53 & 2.65 $\pm$ 0.36 & 2.83 $\pm$ 0.90 & 2.75 $\pm$ 0.63\phantom{$^*$}	  & \bf{2.62 $\pm$ 0.53} \\
\bottomrule        
\end{tabular}
}
\caption{The average RMSE in $^\circ C$ over all weather stations, vineyards, and cultivars per region over five data splits (no transfer). 
Best-in-class results are reported in bold and a $^*$ denotes a statistically significant result $(p<0.05)$ using the paired t-test with respect to the best performing \textit{deployed} model.}

\label{tab:main_results}
\end{table*}

\section{Experiment Setup}
Our proposed cold hardiness model and two modalities of LT50 prediction transfer are evaluated on prediction accuracy compared to the current state-of-the-art baselines. We design our experiments to answer the following questions: 
\begin{enumerate}
\setlength{\itemsep}{0cm}
    \item[\q{1}:] {How does aggregating data across regions compare to the current site- and cultivar- specific modeling paradigm? } 
    \item[\q{2}:] {How do the accuracy of our transfer learning approaches compare to our multi-task model?} 
    \item[\q{3}:] {How does the amount and type of auxiliary information available for approximating $\phi$ influence transfer performance and biological realism? }
\end{enumerate}

\subsubsection{Real World Datasets}
We use six real-world datasets collected from diverse cool-climate viticultural regions in North America. Each region contains one or more weather stations, vineyards, and cultivars. We study six regions: the British Columbia Okanagan Valley (BCOV), the Ontario Niagara Peninsula (ONNP), Washington State (WA), New York State (NY), Michigan State (MI), and Nova Scotia (NS), across 12 cultivars. We use weather data from the nearest weather station that provides daily temperature averages~\citep{c:noaa,c:agnet}. A summary of the datasets used in this study and our data cleaning procedure can be found in \textit{Appendix A}.

\subsubsection{Baselines}
We evaluate two instantiations of our proposed framework: a \textit{Deep Learning (DL)} model and \textit{Hybrid} model. We compare with three deployed LT50 models: (1) \textit{Ferguson Model}--- a biophysical model calibrated on vineyard- and cultivar- specific data~\citep{ferguson2014} and deployed on AgWeatherNet~\citep{c:agnet}; (2) \textit{GrapeHardiNet}--- a multi-task deep learning model trained on vineyard-specific data~\citep{saxena2023a}, deployed on AgWeatherNet~\citep{c:agnet}; and (3) \textit{NYUS 2.2}--- a MLP model~\citep{wang2024a} trained and deployed on NEWA~\cite{c:newa}.

\subsubsection{Model Architecture} 
We choose the widely used Ferguson model~\citep{ferguson2014} as $\mathcal{M}$ for LT50 prediction and implement it in PyTorch~\cite{paszke2017} to enable backpropagation for learning $\psi$ and $\phi$ (details in \textit{Appendix C}). For $\phi$, we learn latent features for each region, weather station, vineyard and cultivar, each of size six, concatenating them with the six weather features as input to $\psi$. For $\psi$, we use a recurrent network consisting of two linear layers, followed by a GRU~\citep{chung2014}, and two more linear layers. We use ReLU activation between linear layers. For the \textit{Hybrid} model, where $\psi$ is used to predict parameters of $\mathcal{M}$, an additional $\tanh$ activation is applied after the final linear layer to scale outputs into the range $[-1,1]$ before rescaling to the varying magnitudes of parameters of $\mathcal{M}$. Details and additional discussion can be found in \textit{Appendix B}.

\begin{table*}[t]
\centering
\setlength{\tabcolsep}{1mm}
{\fontsize{9}{09}\selectfont
\begin{tabular}{l|c|ccc|cc}
\toprule
   & DL Model & Ferguson & Naive Average & Base Embedding ($\phi$)       & Text Descriptions  &  Few shot (5 yr) \\
   \midrule
 
WA & 1.22 & 3.30 & 2.10 & 2.38 & \bf{1.80}$^*$ & 1.99    \\
NY & 1.52 & 2.74 & 1.82 & 2.02 & \bf{1.74}\phantom{$^*$} & 1.78    \\
NS & 1.47 & 2.99 & 1.71 & 2.59 & \bf{1.58}$^*$ & 1.65    \\
MI & 2.75 & 3.17 & 2.90 & 3.19 & \bf{2.63}$^*$ & 2.71    \\
\bottomrule
\end{tabular}
}
\caption{The RMSE in $^\circ$C on four evaluation regions unseen during training for our two methods of transfer across five seeds. We compare against the DL model trained with all regional data, the Ferguson model trained with data from BCOV, naively averaging cultivar predictions from BCOV and ONNP, and the direct embedding from $\phi$. Best-in-region transfer results are reported in bold. A $^*$ denotes a statistically significant improvement with respect to the best-in-region transfer baseline.}
\label{tab:transfer_results}
\end{table*}

\subsubsection{Model Training Protocol}
We split the data into training, validation, and testing sets. In each split, we ensure that there is a representative amount of data for each region, weather station, vineyard, and cultivar. Our testing set consists of two seasons per cultivar per vineyard and our validation set consists of one season per cultivar per vineyard. The remaining data is used for training. We use the validation set for hyperparameter tuning and model selection during training. 

Our models were trained for 400 epochs using a learning rate of 0.0002 and a batch size of 72 seasons using data from all six regions. The Ferguson model was calibrated using stochastic gradient descent for 200 epochs using a learning rate of 0.01 and a batch size of four using only per-cultivar data. The GrapeHardiNet model was trained for 400 epochs using learning rate of 0.002 and a batch size of 12 using per-vineyard data. We used the pretrained NYUS2.2 model which was trained using data from all regions.

For our transfer learning experiments, to simulate predictions for new regions and cultivars, we only used data from the BCOV and ONNP regions for training and used the other four regions for evaluation, or withold one cultivar $c$ across all regions and evalute on $c$. To train our approximators of $\phi$ with text and few seasons of historical data, we froze the weights of our multi-task model $\widehat{\mathcal{F}}$ after 400 epochs and the E5 sentence embedding model. Then, for 200 epochs, we trained the downsampling MLP for our text embeddings jointly with the cross-attention mechanism (using up to five seasons of training data) by randomly selecting a portion of the minibatch to use each latent representation. With this approach, our transfer learning setting requires no model adaptation period as all transfer learning do not require new data for finetuning. Training time averaged two hours per model on an NVIDIA 3080 using the Adam optimizer. 

\subsubsection{Model Evaluation Protocol}
We trained each model five times on different data splits using different seeds and reported the average Root Mean Squared Error (RMSE) across regions, weather stations, vineyards and cultivars. The RMSE is in degrees Celsius for each day in the data sequence that contains a LT50 observation. There are typically five to 30 LT50 observations per season. We compute the RMSE across all seasons and cultivars in a region, or across each cultivar, to obtain a measure of how well each model performs.

\section{Results and Discussion}
\paragraph{Q1: Overall Model Performance} To establish an upper bound on the performance of our transfer learning approaches, we first investigate the performance of our models against the baselines \textit{without transfer} using all regional data. Table~\ref{tab:main_results} shows the average RMSE values for LT50 across all six cool-climate viticultural regions. The results show that both versions of our general model outperform all baselines, offering on average a 9\% improvement over the next best deployed model. Paired t-tests ($p<0.05$) across evaluation seasons confirm that the majority of our improvements over best-in-region deployed model are \textit{statistically significant}. The NYUS2.2 model underperformed across most regions, despite aggregating data from all regions into a single model, indicating the importance of learning expressive $\phi$ and $\psi$. 

\begin{figure}[t]
    \centering
    \includegraphics[width=0.95\linewidth]{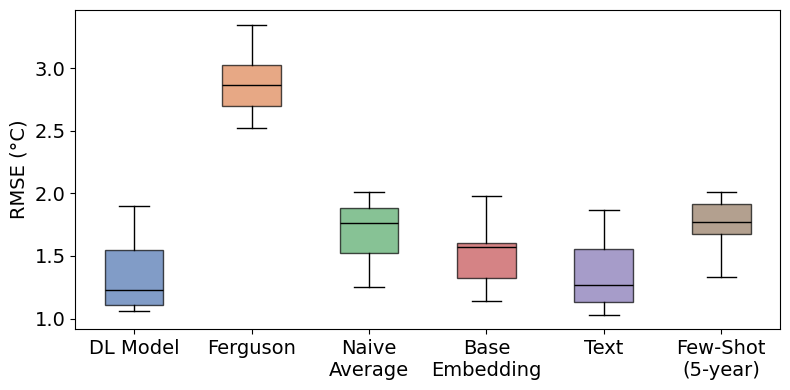}
    \caption{The RMSE across 12 evaluation cultivars, each of which was withheld during training our two methods of transfer across five seeds. We compare against the DL model trained on all cultivars and regions.}
    \label{fig:cultivar_transfer}
\end{figure}

Two key observations from these results are: (i) our DL and Hybrid models outperform the Ferguson model (considered state of the art in viticultural communities) by $1^\circ C$ RMSE in data-limited regions such as New York, Nova Scotia, and Michigan, or equivalently a near 30\% increase in performance; (ii) data aggregation improves performance over the GrapeHardiNet site-specific model, even in regions with many samples (BCOV, ONNP, WA). Together, these results suggest that cross-region data sharing can substantially improve the predictive performance in cold hardiness modeling, challenging the prevailing \textit{site-specific} modeling paradigm.

\paragraph{Q2: Prediction Transfer to Unseen Regions and Cultivars} 

\vspace{3pt}
\noindent \textit{Regional Transfer:~} 
To simulate missing regions and evaluate our transfer learning approaches, we train our DL model using data from only the BCOV and ONNP regions and evaluate the performance on the WA, NY, NS, and MI regions across all 12 cultivars (results for the Hybrid model in \textit{Appendix D}). We compare against four baselines: (i) our DL model trained on all regions to give an upper bound on performance; (ii) the Ferguson model trained on BCOV and ONNP data, as transfer without recalibration is common practice in the viticultural community~\citep{rubio2020}; (iii) naive averaging over the per-cultivar predictions in the BCOV and ONNP regions; and (iv) the base embedding from $\phi$ learned during multi-task training. Each representation for $r,s,v,c$ is sampled from $\mathcal{N}(0,1)$ at initialization, meaning that the region representation may be uninformative if $r$ is not present in the set of training tasks $\mathcal{T}$.

\begin{figure*}[t]
    \centering
    \includegraphics[width=0.90\linewidth]{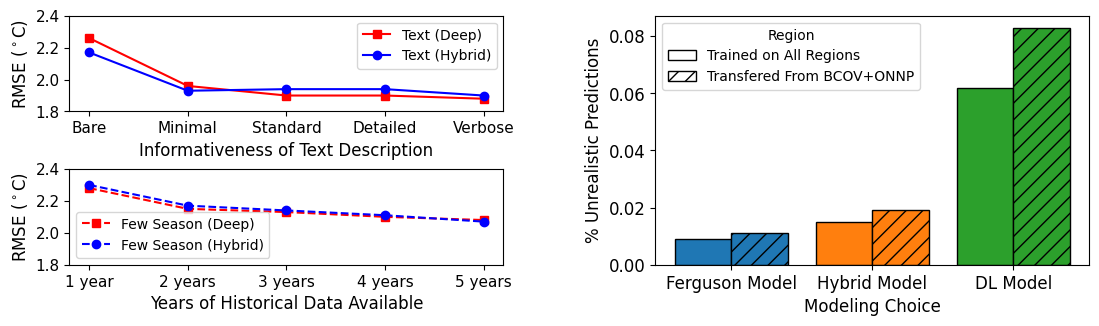}
    \caption{(Left) The RMSE in $^\circ C$ averaged across four regions unseen during training with additional information in the form of more verbose text embeddings or additional years of data for both our approaches. (Right) The percentage of biologically unrealistic predictions over three instantiations of our framework for LT50 prediction. Solid bars are results from models trained on all data, hatched bars are from models trained on BCOV+ONNP data and transferred to other four regions. }

    \label{fig:info_vs_realism}
\end{figure*}

Table~\ref{tab:transfer_results} shows that learning from text outperformed all baselines, achieving comparable performance to the DL model in some regions.
The few-shot approach performed slightly worse despite having access to minimal data from that region during adaptation. We hypothesize this is due to the sparsity and irregularity of samples from some regions, which made it challenging to approximate $\phi$. Unsurprisingly, the base embedding approach performed poorly, as these uninformed regional representations are random noise in $\phi$. Perhaps surprisingly, the naive averaging baseline over cultivars is a strong prior, outperforming the Ferguson model trained on only per-cultivar data in BCOV, indicating the Ferguson model's inability to transfer predictions to new regions. Overall, our methods of transfer provide informed LT50 predictions, serving as an appropriate alternative when data from a new region is not available during training.

\vspace{3pt}
\noindent \textit{Cultivar Transfer:~}
Similar to regional transfer, our approach also enables approximation of latent cultivar features for prediction transfer. We compare against the same baselines as in the Regional Transfer experiment, except that we evaluate our transfer approaches using DL models trained on all regions, withhold one of 12 cultivars as the target for transfer. In Figure~\ref{fig:cultivar_transfer}, we show the distribution of the average RMSE per target cultivar. Our results mirror the findings in our Regional Transfer experiment. The Ferguson model performs poorly under transfer. The naive averaging model is a strong prior, but is not as accurate as using text descriptions. Perhaps surprisingly, the base embedding approach performed comparably, indicating that an uninformative cultivar representation is less impactful than an uninformative regional one on LT50 prediction, which is supported by our previous empirical observations and in \textit{Appendix E}.

\paragraph{Q3: Latent Representation Approximation and Realism}
We now examine how the quantity and informativeness of auxiliary information available for approximating $\phi$ affects the accuracy and biological realism of the predictions.

\vspace{3pt}
\noindent \textit{Prediction Accuracy:~} 
We conducted an ablation study to understand the information required for successful transfer. Specifically, we construct four additional text embeddings ranging from Bare (the name of the cultivar/region) to Verbose (the name of the cultivar/region, latitude/longitude, key traits relating to viticulture and cold hardiness; examples in \textit{Appendix D}) and use our few shot cross-attention mechanism averaging over one to five years of data. Figure~\ref{fig:info_vs_realism} (left) demonstrates that increased information helps learn a potentially more useful latent representation and reduces the resulting RMSE, averaged across the four evaluation regions. Interestingly, both the DL and Hybrid models perform similarly, suggesting that once sufficiently informative latent representations are available, the biophysical model $\mathcal{M}$ contributes little additional predictive value or regularization.

\vspace{3pt}
\noindent \textit{Biological Realism:~}
Biologically realistic predictions are vital for deployment and adaption by growers in the field. In consultation with domain experts, we define a prediction as biologically \emph{unrealistic} if it satisfies either of the two criteria: (i) the predicted LT50 value never reaches the known minimum LT50 for the cultivar ($-10$ or $-5$ $^\circ C$) during spring deacclimation and (ii) the predicted LT50 decreases after reaching maximum value despite exposure to warming spring temperatures. If a day in the season meets either of these criteria, it is marked as biologically unrealistic, and we report the percentage of unrealistic predictions during a season. The DL model exhibits both phenomena, resulting in confusing predictions that could mislead growers. 

Evaluating our general modeling framework across all six regions with and without transfer approaches (Figure~\ref{fig:info_vs_realism}, right) demonstrates that the DL model has a substantially higher rate of biologically unrealistic predictions compared to the Hybrid model (6.2\% vs 1.9\% across all cultivars). Interestingly, it also has a larger increase in unrealistic predictions in the transfer setting, indicating how the biophysical model $\mathcal{M}$ can aid prediction realism in out-of-distribution evaluation settings. While small, the differences have influenced the choice to \textit{deploy} the Hybrid model compared to the DL mode, in collaboration with our project partners, despite its slightly lower accuracy. 

\section{Summary and Future Work}
Current cold hardiness models hosted on platforms, such as AgWeatherNet and NEWA~\citep{saxena2023a,wang2024a}, are unable to make predictions for cultivars and regions outside of the available training data. In contrast, our transfer learning methodology provides multiple ways of transferring predictions to unseen regions and cultivars with limited or no prior data, and our framework enables systematic comparison between modeling paradigms. Our Hybrid model is currently deployed on AgWeatherNet for the `26-`27 season and we are actively exploring deployment of our transfer learning approach to aid decision support for regions with limited data. While we focus on the cold hardiness domain, motivated by the operational needs of our project partners and public data availability, our framework is applicable to a wide range of prediction tasks, specifically crop state tasks where data is scarce across cultivars and farms. Future work will investigate extending our approach to cherries, blueberries, and apples, key crops in the Pacific Northwest. 

\clearpage

\section*{Acknowledgments}
The authors thank Lynn Mills and Zilia Khaliullina at the Irrigated Agriculture Research and Extension Center (IAREC), Washington State University, for their invaluable support in collecting and sharing grape cold hardiness data. This research was supported by USDA NIFA award No. 2021-67021-35344 (AgAID AI Institute). 

\bibliography{ref}

\clearpage

\section{Appendix A: Dataset Description and Processing}
\subsection{Appendix A.1: Real World Dataset}
As mentioned in the main text, our experiments are conducted on six real world datasets from six diverse cool-climate viticultural regions (British Columbia Okanagan Valley, Ontario Niagara Peninsula, Washington State, New York State, Nova Scotia, and Michigan). We choose these regions for their available data that either collected in collaboration with our project partners or publicly available for download, and abundance of vineyards, cultivars and seasons of observations. The cold hardiness of 12 cultivars (Cabernet Franc, Cabernet Sauvignon, Chardonnay, Gewurztraminer, Merlot, Pinot Blanc, Pinot Gris, Pinot Noir, Riesling, Sauvignon Blanc, Syrah, and Zinfandel) has been measured in these regions using Differential Thermal Analysis (DTA)~\cite{mills2006} by collecting cane samples from dormant buds at daily, weekly, biweekly, or monthly intervals from September 7th to May 15th (the dormancy season). Table~\ref{tab:data} shows the distribution of LT50 samples across the maximum number of years in the regional dataset along with the number of weather stations, vineyards, and cultivars present in each region. A majority of the data is publicly available for download\footnote{https://github.com/imbaterry11/NYUS.2} while additional data may be made available by our project partners upon request. 

In addition to the LT50 observations, we use real-world weather data recorded daily from nearby open-field weather stations. These weather features include the daily minimum, maximum, and average temperatures and the daily precipitation. To obtain this data, we use a combination of AgWeatherNet~\cite{c:agnet} weather stations, Canadian government weather stations\footnote{https://climate.weather.gc.ca/}, and NOAA weather stations\footnote{https://www.ncei.noaa.gov/access/past-weather/}. We record the latitude and longitude of each vineyard from which the LT50 samples were collected and choose the nearest weather station that has historical daily data for all seasons for which there are recorded LT50 observations. 

\subsection{Appendix A.2 Data Progressing Procedure}
Historical weather data is noisy and sometimes contains missing weather observations. To make the data usable, we process it in two ways: (1) If any of the temperature features are missing morethan 10\% of the values for the season, we discard the entire season along with the LT50 values from training. Otherwise, we linearly interpolate between the two nearest observed values. (2) We normalize all weather features using z-score normalization. We include a sinusoidal embedding for the day of the season using the equation $(x_1,x_2)=[\sin(\mathrm{doy} \bmod{365}), \cos(\mathrm{doy} \bmod{365})\big]$. We include any season with at least one valid LT50 observation. Days with missing LT50 observations are recorded as NaN and masked during loss computation and backpropagation. 

Inherently, LT50 samples from DTA are noisy~\cite{ferguson2011} due to the process of freezing multiple dormant buds at the same time. Moreover, different viticultural labs across each region may use a slightly different DTA process which we cannot directly account for and could account for the larger standard deviations recorded in the results section. We believe that part of the benefit of learning regional specific latent features can address some of the variability observed in LT50 samples amount the same cultivars across regions. 

\begin{table}[t]
\centering
\setlength{\tabcolsep}{1.5mm}
{\fontsize{9}{09}\selectfont
\begin{tabular}{lrrrrr}
\toprule
Regions & Stations & Vineyards & Cultivars & Years  &  \# Samples      \\
\midrule
BCOV    & 4 & 11 & 12 &  9 & 3073 \\
ONNP    & 2 & 10 & 12 & 11 & 4720 \\
WA      & 1 & 1  & 12 & 27 & 3547 \\
NY      & 4 & 4  & 11 & 5  & 2373 \\
NS      & 1 & 1  &  3 & 6  &  313 \\
MI      & 1 & 1  &  4 & 4  &  404 \\  
\bottomrule     
\end{tabular}
\caption{The distribution of weather stations, vineyards, cultivars, maximum number of years of data and total LT50 observations across the ten cool-climate viticultural regions used in this study.}
\label{tab:data}
}
\end{table}

\begin{figure*}[t]
    \centering
    \includegraphics[width=0.8\linewidth]{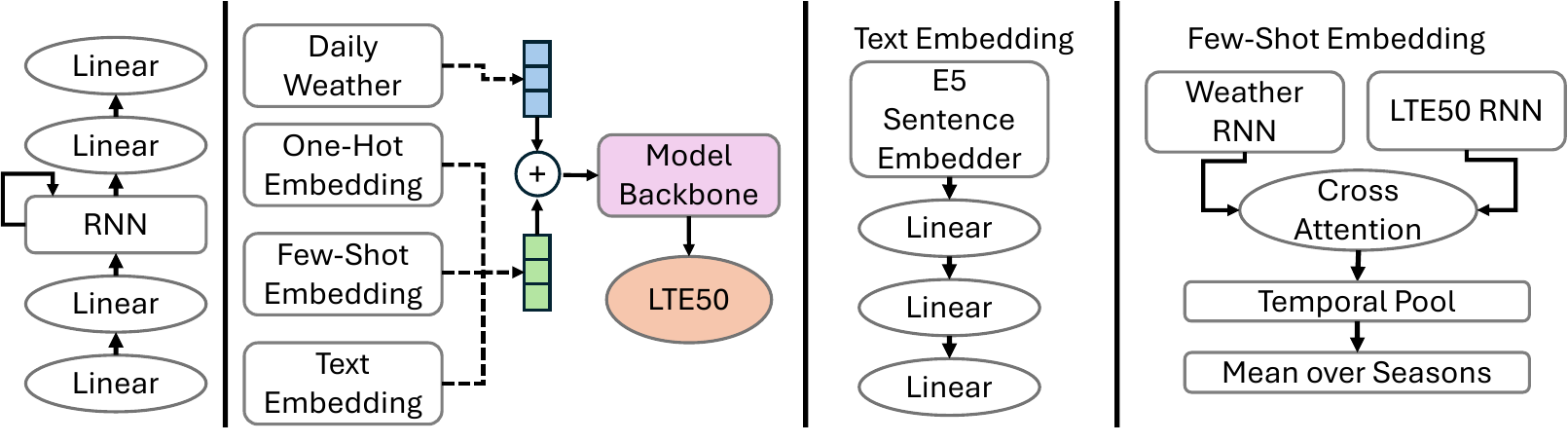}
    \caption{An overview of our modeling framework with the model backbone ($\psi$, far left) and the interaction between $\phi$, approximations of $\phi$, and $\psi$ (middle left). The text embedding model architecture (middle right) and the few shot embedding approach with cross-attention (far right) comprise of our transfer learning framework. }
    \label{fig:transfer_overview}
\end{figure*}

\section{Appendix B: Model and Hyperparameter Selection}
\subsection{Appendix B.1: Evaluation Protocol}
We outline additional portions of our experimental protocol for the purpose of reproducibility. All experiments were run on an Ubuntu 24.04 system with a NVIDIA 3080 GPU with 10GB of VRAM. As noted in the main text and \textit{Appendix A}, there is high variance in measured LT50 across cultivars. To address this challenge in validation and evaluation, we build our training and evaluation set as follows. 

For five different seeds, we first withhold two seasons of data per cultivar per vineyard (and region) for the testing set. We then withhold an additional season of data per cultivar from the training data. For the most data scarce cultivars and regions, this results in one season of data for the training set and one season of data for the validation set (Michigan). Otherwise, we have at least two seasons of data per cultivar in the training set. At most, we have 24 years of data in the training set for the Riesling cultivar in the WA region. 

Our transfer experiments used the BCOV and ONNP regions as training regions, despite WA and NY being viable candidates. We chose to do so given the distribution of vineyards in these regions. Each weather station in the region has one or more associated vineyards. Therefore, we can think of these regions as the most diverse in their coverage, as any model will see representations for multiple vineyards and weather stations within the training data compared to other regions. Thus, we choose to evaluate the WA, NY, NS, and MI regions as sources of transfer. It should be noted that the WA and NY regions are well studied, but do not have the same coverage of cold hardiness observations from multiple weather stations, or geographically dense vineyards, which would have impacted the evaluation of our transfer learning experiments.   

\subsection{Appendix B.2: Model Architecture}
In this section, we describe our modeling architecture choice for both the Deep Learning and Hybrid models. In order to motivate our choice of architecture, we draw on previous work from both the viticultural and artificial intelligence (AI) communities. \citet{badeck2004} noted the complex temporal relationship between weather and crop physiology with respect to early spring budding and cold hardiness. This observation motivates our choice of $\psi$, the function approximator in our framework. While the NYUS2.2 model~\citep{wang2024a} uses a multi-layer perceptron, the accuracy of their model is a result of extensive feature engineering, using the exponential weighted average of up to 20 days of previous weather. Instead, we follow~\citet{saxena2023a} and use a recurrent network, specifically a GRU~\cite{chung2014}. We augment the GRU with two linear layers before and after, with ReLU activation between the linear layers. These layers serve to scale the input up to the hidden dimension of the GRU and down to the size of the output. We considered more expressive function approximators (e.g. Transformers~\citep{vaswani2017}), but found empirically that they did not improve the performance of our models while adding substantially more parameters and training time. 

To provide latent information to the model, we choose to concatenate the learned representations for the cultivar, vineyard, weather station, and region together along with the processed daily weather features before passing them into the first linear layer of our model backbone (see Figure~\ref{fig:transfer_overview}) for daily LT50 prediction. Prior work~\citep{solow2026} has shown that concatenation is the most efficient form of data sharing for latent representations in these crop prediction domains, and so we adopt their concatenation approach and extend it to include the region, weather station, and vineyard representations. To encourage equal influence of the region, weather station, vineyard, cultivar, and weather on the LT50 predictions, we let the size of each learned embedding for $r$, $s$, $v$ and $c$ be the same size as the number of weather features. In the case of our experiments, we have six weather features, resulting in a total embedding size of $n=24$. Thus, the input to the first linear layer of $\phi$ is 30-dimensional. The first linear layer is $1/4$ the size of the GRU and the second is $1/2$ the size, as with the layers proceeding the GRU. 

In our framework the function approximator $\psi$ can either output daily LT50 predictions or daily parameter predictions of a biophysical model. In the first case, we include no additional activation function after the last linear layer, letting it instead directly predict the cold hardiness. In the hybrid modeling case, we add an additional $\tanh$ activation function after the last linear layer to scale the output into the range $[-1,1]$. We found that this approach is better than predicting daily model parameters directly as the ranges of the parameters have varying magnitudes ($10^{-1}$ to $10^2$) which makes direct parameter prediction difficult. Preselected ranges for each parameter in the Ferguson model can be found in Table~\ref{tab:model_params}.
\begin{table}[t]
\centering
\setlength{\tabcolsep}{1mm}
{\fontsize{9}{09}\selectfont
\begin{tabular}{lrrrr}
\toprule
           \makecell{Models}      & \makecell{Hidden\\Size} & \makecell{Learning\\Rate} & \makecell{Batch\\Size} & \makecell{Learning Rate\\Anneal} \\
\midrule
Deep Learning         & 2048                            & 0.0001                            & 72                             & 0.9                           \\
Hybrid          & 1024                            & 0.0002                             & 72                              & 0.9                           \\
GrapeHardiNet              & 2048                            & 0.0001                            & 12                             & 0.9                           \\
                     
\makecell{Ferguson  w/ SGD} & N/A         & 0.1                               & 4                              & 0.9                           \\
\bottomrule
\end{tabular}
}
\caption{Best hyperparameters found for each model type after five-fold cross validation on the entire LT50 dataset. SGD: Stochastic Gradient Descent.}
\label{tab:hyperparameters}
\end{table}

\begin{table*}[t]
\centering
\setlength{\tabcolsep}{1mm}
{\fontsize{9}{09}\selectfont
\begin{tabular}{llllrr}
\toprule
 Parameter Name & \multicolumn{2}{l}{Parameter Description}                          & Unit                   & \multicolumn{1}{l}{Min Value} & \multicolumn{1}{l}{Max Value} \\        \\
 \midrule
 HCINIT         & Initial Cold-Hardiness                               &             & $^\circ C$             & -15                           & 5                             \\
 HCMIN          & Minimum Cold-Hardiness                               &             & $^\circ C$             & -5                            & 0                             \\
 HCMAX          & Maximum Cold-Hardiness                               &             & $^\circ C$             & -40                           & -20                           \\
 TENDO          & Base Temperature During Endodormancy                 &             & $^\circ  C$            & 0                             & 10                            \\
 TECO           & Base Temperature During Ecodormancy                  &             & $^\circ C$             & 0                             & 10                            \\
 ENACCLIM       & Acclimation Rate During Endodormancy                 &             & $^\circ C^\circ C^{-1}$ & 0.2                           & 0.2                           \\
 ECACCLIM       & Acclimation Rate During Ecodormancy                  &             & $^\circ C^\circ C^{-1}$ & 0.2                           & 0.2                           \\
 ENDEACCLIM     & Deacclimation Rate During Endodormancy               &             & $^\circ C^\circ C^{-1}$ & 0.2                           & 0.2                           \\
 ECDEACCLIM     & Deacclimation Rate During Ecodormancy                &             & $^\circ C^\circ C^{-1}$ & 0.2                           & 0.2                           \\
 ECOBOUND       & Threshold for Ecodormancy Transition                 &             & $^\circ C$             & -800                          & -200                          \\
\bottomrule
\end{tabular}
}
\caption{The parameters of the Ferguson Model used in our hybrid approach. The ranges correspond to the minimum and maximum values that the parameter can be after $\tanh$ activation normalizing from the range $[-1,1]$}
\label{tab:model_params}
\end{table*}

\subsection{Appendix B.3: Hyperparameter Selection}
To select a hyperparameter set for our model evaluations, we consider four different hyperparameters for our 5-fold validation. 1) Number of GRU hidden units in $[128, 256, 512, 1024, 2048]$, (2) Learning rate in $[0.01, 0.005, 0.001, 0.0005, 0.0002, 0.0001]$, (3) Batch size in $[4, 8, 12, 16, 24, 36, 72, 96]$, (4) Learning rate annealing in $[0.8, 0.85, 0.9, 0.95, 1]$ We also considered the omission of the extra linear layers before the GRU and considered Transformer architecture before. 

We performed a grid search over these parameters for the DL and Hybrid models, and for the Ferguson and GrapeHardiNet models which we trained from scratch. We did not train the NYUS2.2 model and instead downloaded it from a publicly available checkpoint~\citep{wang2024a}. Our experiments in the main text use varying amounts of data. We found empirically that varying these hyperparameters did not meaningfully impact the performance as we excluded regions from training during our transfer learning experiments or excluded learned latent representations for the region, weather station, or vineyard. The best hyperparameters based on the performance on the validation set are in Table~\ref{tab:hyperparameters}. The difference in batch size is a direct reflection on the scale of data that the models are trained on. The Ferguson model is always trained on cultivar specific data. GrapeHardiNet is always trained on vineyard specific data. Meanwhile, the Deep Learning and Hybrid model are trained on data across all regions. 

\section{Appendix C: Ferguson Biophysical Model}
Cold hardiness is a property of dormant buds of woody plants. The Lethal Temperature Exotherm (LTE) characterizes the temperature at which $X\%$ of a plant's dormant buds freeze. The most common measurement is the LT50 value, or the temperature at which $50\%$ of dormant grapevine buds freeze. LTE10 and LTE90 measurements are recorded across some datasets, but LT50 is the most commonly used value reported by models and used for decision support in the field. LT50 is typically measured and reported from September 7th to May 15th of each year in the Northern hemisphere in cool climate viticultural regions. Cold hardiness is difficult to measure in the field and consequently growers rely on models to inform their decision making during the fall, winter, and spring months.

The Ferguson model~\citep{ferguson2011,ferguson2014} provides daily LTE10, LT50, and LTE90 predictions, calibrated based on data in the Washington region. By contrasting the predictions of the Ferguson model with the weather forecast, grape gowers decide whether preventative measures (e.g. deploying wind machines or kerosene heaters in the field) are needed to protect dormant buds from a catastrophic freezing event. The Ferguson model computes the change in LT50 as a function of daily acclimation and deacclimation based on the dormancy stage and daily average temperpature. See~\citet{ferguson2011} for a complete description of the mathematical model. The Ferguson model parameters that we calibrate in our hybrid approach are listed in Table~\ref{tab:model_params}.

\section{Appendix D: Transfer Learning Representations}
Section 4 in the paper describes our two approaches for prediction transfer to approximate the latent representation function $\phi$: using (i) text descriptions and (ii) limited historical data. In this section we provide additional details on the information and data used to train our two methods of transfer. 

\subsection{Transfer via Text}
Textual descriptions are a powerful means of encoding domain specific information, especially to non-experts. In conjunction with viticultural experts on our team, we identified key attributes of cultivars and regions that could be encoded in text. Using these attributes, we created sentence queries for each cultivar and region that could be passed to any language model to produce an embedding. 

Given that cultivars with similar attributes should have similar LT50 behaviors, we need a model that encodes semantically similar sentences to the same regions in the embedding space. Sentence embedding models are a natural tool for this, and we use the recent E5 sentence embedding model~\cite{wang2024multilingual} for our experiments given its high rating on semantic search and information retrieval tasks. Below, we list all the text generated for the cultivars and regions. In our anonymized code (provided in supplemental materials), the text strings for the weather stations and vineyards can be found as well:

\subsubsection{Cultivar Descriptions}
\begin{enumerate}
    \item  Cabernet Franc is an early- to mid-budding, mid-ripening red grape cultivar with medium vigor and loose to moderately compact clusters, producing wines with moderate acidity and a lighter, translucent red color. It has good cold-hardiness (tolerating around -15C/5F) and prefers well-drained soils such as gravelly loam or limestone.
    \item Cabernet Sauvignon is a late-budding, late-ripening red grape cultivar with small, compact clusters, producing deeply colored wines with firm acidity and strong tannins. It has moderate cold-hardiness (tolerating around -15C/5F) and prefers well-drained gravelly or sandy-loam soils.
    \item  Chardonnay is an early- to mid-ripening white grape cultivar with moderate vigor and medium-sized, loosely packed clusters, producing wines with bright acidity and a pale to golden color. It has moderate cold-hardiness (tolerating around -15C/5F) and prefers well-drained soils such as limestone, chalk, or clay.
    \item Gewurztraminer is an early- to mid-ripening white grape cultivar with low to moderate vigor and tightly packed clusters, producing intensely aromatic wines with low to moderate acidity and a golden to deep yellow color. It has low to moderate cold-hardiness (tolerating around -12C/10F) and prefers well-drained, calcareous or loamy soils.
    \item Merlot is a mid-budding, mid- to late-ripening red grape cultivar with medium vigor and medium-sized, loosely packed clusters, producing wines with soft tannins, moderate acidity, and a deep ruby color. It has moderate cold-hardiness (tolerating around -15C/5F) and prefers well-drained clay or gravelly soils.
    \item Pinot Blanc is an early- to mid-ripening white grape cultivar with moderate vigor and medium-sized, loosely packed clusters, producing wines with moderate acidity and a pale yellow to straw color. It has moderate cold-hardiness (tolerating around -15C/5F) and prefers well-drained, calcareous or loamy soils.
    \item Pinot Gris is an early- to mid-ripening white grape cultivar with moderate vigor and medium-sized, loosely packed clusters, producing wines with medium acidity and a color ranging from pale yellow to coppery gray. It has moderate cold-hardiness (tolerating around -15C/5F) and prefers well-drained, fertile soils such as loam or limestone.
    \item   Pinot Noir is an early-budding, mid- to late-ripening red grape cultivar with low to moderate vigor and small, loosely packed clusters, producing wines with bright acidity and a light to medium ruby color. It has moderate cold-hardiness (tolerating around -15C/5F) and prefers well-drained, calcareous or loamy soils.
    \item   Riesling is an early- to mid-budding, late-ripening white grape cultivar with moderate vigor and small to medium, compact clusters, producing wines with high acidity and a pale yellow to golden color. It has good cold-hardiness (tolerating around -20C/-4F) and prefers well-drained soils such as slate, limestone, or loam.
    \item Sauvignon Blanc is an early- to mid-ripening white grape cultivar with moderate vigor and medium-sized, loose clusters, producing wines with high acidity and a pale greenish-yellow color. It has moderate cold-hardiness (tolerating around -15C/5F) and prefers well-drained soils such as gravel, limestone, or sandy loam.
    \item   Syrah is a mid- to late-budding, mid- to late-ripening red grape cultivar with high vigor and medium to large, compact clusters, producing wines with moderate acidity and a deep, dark purple color. It has moderate cold-hardiness (tolerating around -15C/5F) and prefers well-drained, stony or gravelly soils.
    \item  Zinfandel is a mid- to late-budding, mid- to late-ripening red grape cultivar with high vigor and medium to large, loosely packed clusters, producing wines with moderate acidity and a deep ruby to purple color. It has low to moderate cold-hardiness (tolerating around -12C/10F) and prefers well-drained, sandy or loamy soils.
\end{enumerate}

\subsubsection{Regions}

\begin{enumerate}
    \item The Okanagan Valley in British Columbia is a semi-arid, continental-influenced wine region with long, warm, sunny summers, cool nights, and low annual precipitation, where well-drained soils and pronounced diurnal temperature swings create ideal conditions for ripening a wide range of vinifera grapes while preserving acidity and flavor concentration.

    \item The Ontario Niagara Peninsula is a cool-climate, lake-moderated wine region with warm, sunny summers, cool nights, and fertile, well-drained soils over limestone and glacial deposits, creating an extended growing season that supports balanced ripening and high-quality vinifera grape production.

    \item Washington State is a semi-arid, continental-influenced wine region with hot, sunny days, cool nights, and low annual precipitation in the rain-shadow of the Cascade Mountains, where well-drained alluvial and volcanic soils combined with irrigation allow precise control of vine vigor and high-quality vinifera grape production.

    \item New York State in the United States is a cool- to moderate-climate wine region strongly influenced by large lakes and river valleys, with warm summers, cool nights, and cold winters moderated locally by the Finger Lakes, Lake Erie, and Lake Ontario, where glacially derived soils of shale, silt, and gravel support balanced vine growth and high-quality vinifera grape production.

    \item Nova Scotia in Canada is a cool maritime wine region influenced by the Atlantic Ocean and the Bay of Fundy, with mild summers, cool nights, and relatively moderate winters, where well-drained glacial and coastal soils and a long, cool growing season support the production of fresh, aromatic vinifera and hybrid grape varieties.

    \item Michigan in the United States is a cool-climate, lake-moderated wine region surrounding Lake Michigan where warm summer days, cool nights, and long autumns extend the growing season, and well-drained glacial sandy loams and gravelly soils support balanced vine growth and high-quality vinifera grape production.
\end{enumerate}

\begin{figure}[t]
    \centering
    \includegraphics[width=\linewidth]{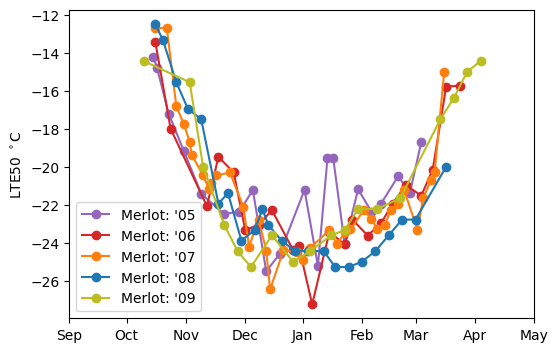}
    \caption{Five years of LT50 samples collected for the Merlot cultivar in the Washington region. Samples are collected aperiodically, with collection starting at different points in the season, and contain high variance.}
    \label{fig:few_seasons}
\end{figure}

\begin{figure}[t]
    \centering
    \includegraphics[width=0.9\linewidth]{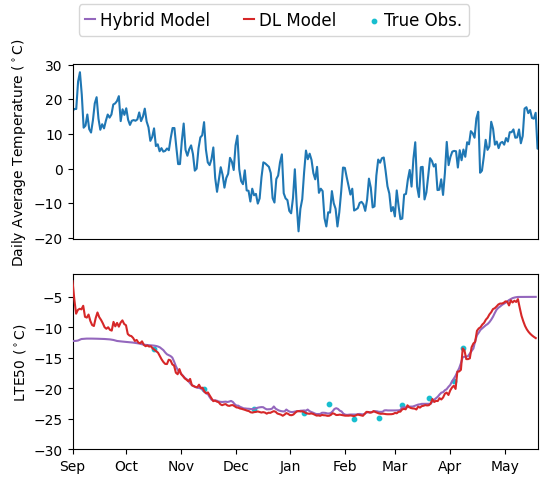}
    \caption{Predictions of our DL model and Hybrid model, and the observed LT50 values of the Chardonnay cultivar at the Roza.2 site in Washington, USA.}
    \label{fig:biological_realism}
\end{figure}

\subsection{Transfer via Few Seasons of Data}
While text transfer enables zero-shot transfer to new regions without prior data, it is also possible to approximate $\phi$ for a new region and cultivar from historical data. However, this is challenging given the nature of historical LT50 observations: samples are not always periodically collected and contain large variance. Moreover, they are conditioned on the weather which adds an additional source of noise. Figure~\ref{fig:few_seasons} shows historical LT50 observations for the Merlot cultivar for the dormant seasons from 2005-2006 to 2009-2010 collected from the Washington region at a single research vineyard. Observe that even at a research vineyard, samples collected by expert viticulturists are sometimes aperiodic. Thus, encoding these observations into a reliable embedding $\phi$ is challenging. Moreover, the sampling rate observed in datasets may vary from region to region, making transfer more challenging when a new sampling period is observed. 

This variance in the data may explain why transfer via text outperformed transfer via few seasons of data. We would expect that transfer from few seasons of data to perform better than text, as it is a few shot method as opposed to a zero shot method. However, our results show that the text embeddings provide a very strong prior over $\phi$ for predictive performance in our multi-task model.

\section{Appendix E: Additional Experiments}
In this section we address (1) the biological realism of cold hardiness models, (2) further understanding the latent space of $\phi$, and (3) the hybrid model transfer results. 

\paragraph{Biological Realism}
As discussed in the main text, the primary concern with using deep learning models for cold hardiness prediction is that they may produce biologically unrealistic predictions that make farmers act on poorly informed management decisions. In Figure~\ref{fig:biological_realism} we show the daily average temperature for a single season with the predictions of our DL model and Hybrid model. Building upon our discussion in the main text, observe that the DL model makes abnormally high predictions in the early season and its prediction curve drops off unexpectedly in the late season. Both of these times are critical for accurate LT50 prediction where the cold hardiness of dormant buds is changing rapidly with the warming or cooling temperatures. 

In particular, despite a warming trend in the late season, the LT50 prediction of the DL model model drops off unexpectedly. In contrast, the  Hybrid model does not exhibit abnormal early- or late-season behavior, instead producing biologically realistic predictions for the entirety of the growing season. As a result, the hybrid model has been \textit{deployed} despite the DL model exhibiting slightly higher testing accuracy.

\begin{table}[t]
\centering
\setlength{\tabcolsep}{1mm}
\renewcommand{\arraystretch}{1.5}
{
\resizebox{\columnwidth}{!}{%
\Large
\begin{tabular}{l|rrrr|rrr|rr|c}
\toprule
     & \multicolumn{4}{c}{Cult. Emb.}                            & \multicolumn{3}{c}{Vine. Emb.}                 & \multicolumn{2}{c}{Stat. Emb.} & \multicolumn{1}{c}{DL Model}   \\
     \hline
     & Cult.          & Vine.          & Stat.       & Reg.        & Vine.          & Stat.       & Reg.        & Stat.         & Reg.         & Reg.        \\
     \midrule
BCOV & 1.31 & 1.25 & 1.25 & 1.34 & 1.21 & 1.19 & 1.39 & 1.18   & 1.28  & 1.13 \\
ONNP & 1.19 & 1.09 & 1.06 & 1.13 & 1.08 & 1.07 & 1.19 & 1.06   & 1.11  & 1.01 \\
\bottomrule
\end{tabular}}
}
\caption{Average RMSE ($^\circ C$) of the DL model using learned latent representations at different levels of data aggregation for BCOV and ONNP regions. Cultivar, vineyard, station, and region denote embeddings learned from data aggregated at the corresponding geographic scale.}

\label{tab:embedding_results}
\end{table}

\paragraph{2a: Data Aggregation Under Varying $\phi$}
The primary focus of our modeling approach relies on learning latent features for the region, weather station, vineyard, and cultivar for data sharing and downstream transfer. We choose to concatenate these features together as input to $\psi$ in along with the weather as prior work has shown the efficacy of this approach for cold hardiness prediction~\citep{solow2026}. However as prior work~\cite{wang2024a} has aggregated data \textit{without} learning regional features, we investigate if these latent features actually improve performance in the presence of a nonlinear function approximator $\psi$. To do so, we consider three variations of our DL model: a Cultivar Embedding model, where the latent space is conditioned soley on the cultivar, a Vineyard Embedding model, where the latent space is conditioned on the cultivar and vineyard, and a Station Embedding model, where the latent space is conditioned on the cultivar, vineyard, and weather station. These three models are compared against the DL model, which latent space is defined by $\phi$, a mapping from cultivars, vineyards, weather stations, and regions to $\mathbb{R}^n$.

In addition, we train these models on varying degrees of data aggregation. For the Cultivar Embedding model, we train only on data aggregated across cultivars at a single vineyard. We then train this model on all vineyards in a region and report the performance. We also train this model on data aggregated across vineyards associated with a single weather station, all vineyards in a region, and all vineyards in the dataset, forming the Cult., Vine., Stat., and Reg., columns in Table~\ref{tab:embedding_results}. We likewise do the same for the Vineyard Embedding and Station Embedding models. There is no difference in data aggregation for our DL model, as we already consider all latent features aggregated across all available regions. We compare across the two principle regions that we train our models on before transfer, BCOV and ONNP given their distribution of cultivars, vineyards, and weather stations throughout each region. 

Table~\ref{tab:embedding_results} shows the average RMSE across the BCOV and ONNP regions comparing different variations of our DL model. As additional data from weather stations or regions is added, performance generally decreases without the addition of complementary latent features (e.g. the addition of multi-region data without learned regional latent features). The best performance is achieved when all latent features and data are used (the DL model), demonstrating the importance of data aggregation with additional latent features for efficient information sharing. 

\begin{figure}[t]
    \centering
    \includegraphics[width=0.8\linewidth]{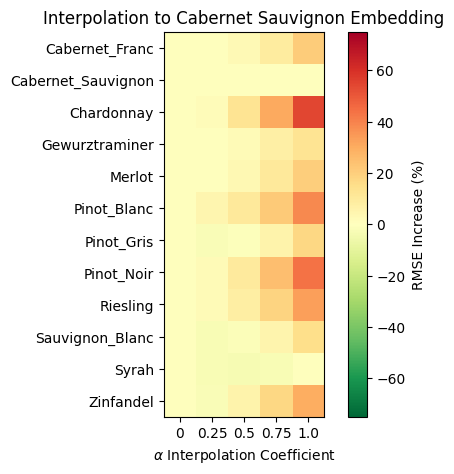}
    \caption{The \% RMSE ($^\circ C$) increase of all 12 cultivars across six regions our DL model, linearly interpolating between the learned cultivar embedding and the Cabernet Sauvignon embedding.  }
    \label{fig:cab_sauv_interpolation}
\end{figure}

\paragraph{2b: Interpolation of $\phi$}
In our multi-task model, we introduced $\phi$ as a function that learns the latent representation for a region, weather station, vineyard and cultivar to provide additional information to weather for accurate prediction across regions. We built intuition in the main text by showing how varying the learned representations $\phi$ across all cultivars resulted in different predictions. However, given that our transfer approach is built off of approximating $\phi$, it is interesting to understand how varying the learned latent representation affects performance accuracy. 

To simulate this, we consider a linear interpolation between the source embedding for a cultivar $c$ and a target cultivar $c'$. As an example, we choose $c'$ to be the Cabernet Sauvignon cultivar as in Figure~\ref{fig:cab_sauv_interpolation}. For every other cultivar $c$, we linearly interpolate with interpolation coefficient $\alpha$ between the source and target embeddings and evaluate on the source cultivar data across all six regions. We report the percent increase in RMSE ($^\circ C$). For the majority of the cultivars, the accuracy degrades consistently as the embedding trends towards $c'$, resulting in a 20\% to 40\$ increase in accuracy (e.g. if the error was $1.0^\circ C$ RMSE, the result of using $c'$ would be $1.40^\circ C$. From this, we can conclude that the individual learned latent representations from $\phi$ have direct impact on the accuracy of the model. Accurately approximating $\phi$ in our transfer learning approaches is crucial for downstream prediction accuracy. 

\begin{figure}
    \centering
    \includegraphics[width=\linewidth]{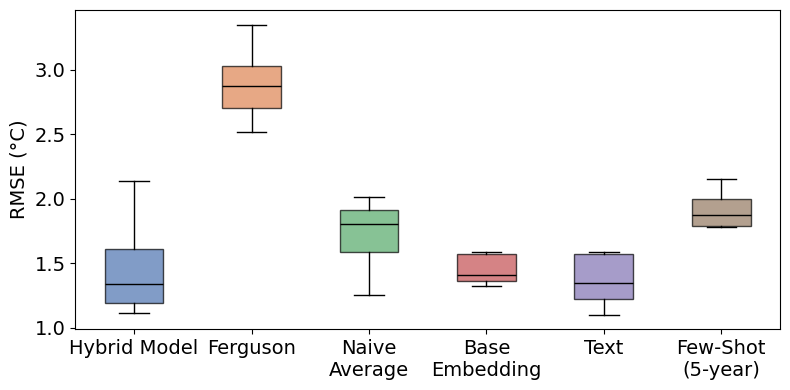}
    \caption{The RMSE across 12 evaluation cultivars, each of which was withheld during training our two methods of transfer across five seeds. We compare against the Hybrid model trained on all cultivars and regions.}
    \label{fig:cultivar_transfer}
\end{figure}

\begin{table*}[t]
\centering
\setlength{\tabcolsep}{1mm}
{\fontsize{9}{09}\selectfont
\begin{tabular}{l|c|ccc|cc}
\toprule
   & Hybrid Model & Ferguson & Naive Average & Base Embedding ($\phi$)       & Text Descriptions  &  Few shot (5 yr) \\
   \midrule
WA & 1.23 & 3.30 & 2.15 & 2.53 & \bf{1.89}$^*$ & 2.27    \\
NY & 1.63 & 2.74 & 1.91 & 2.13 & \bf{1.85}\phantom{$^*$} & 2.30    \\
NS & 1.53 & 2.99 & 1.85 & 2.01 & \bf{1.75}$^*$ & 1.93    \\
MI & 2.62 & 3.17 & 2.93 & 2.76 & \bf{2.76}$^*$ & 2.55    \\
\bottomrule
\end{tabular}
}
\caption{The RMSE in $^\circ$C on four evaluation regions unseen during training for our two methods of transfer across five seeds. We compare against the Hybrid model trained with all regional data, the Ferguson model trained with data from BCOV, naively averaging cultivar predictions from BCOV and ONNP, and the direct embedding from $\phi$. Best-in-region transfer results are reported in bold. A $^*$ denotes a statistically significant improvement with respect to the best-in-region transfer baseline.}
\label{tab:regional_transfer_results}
\end{table*}

\paragraph{Hybrid Model Transfer}
In the main paper, we showed transfer results the results from the DL model in transferring to previously unseen regions and cultivars. We showed the DL regions in the main paper to establish an upper bound on performance, even though our Hybrid model is the one that has been deployed. We repeat the experiments in the main paper using the Hybrid model instead of the DL model and show the results on regional transfer in Table~\ref{tab:regional_transfer_results} and the results on cultivar transfer in Figure~\ref{fig:cultivar_transfer}. 

In line with the results from the DL model, the text description method of transfer outperformed all other approaches and baselines. As we observed in the main paper, the Hybrid model performs slightly worse than the DL model, and the results are shifted accordingly. Interestingly, the Base Embedding baseline did not perform as poorly in the Hybrid model. We hypothesize that the regularization of the biophysical model $\mathcal{M}$ may help with this. Furthermore, the gap between the best performance with the Hybrid model and the text description baseline was less pronounced compared to the DL model. Overall, these results confirm our findings in the main paper, and benchmark the Hybrid model's transfer performance. 

These results demonstrate the utility of our multi-task framework, the importance of the latent embeddings, and the performance of the Hybrid model in the transfer setting. 



\end{document}